\documentclass[letterpaper, 10 pt, conference]{ieeeconf}  

\IEEEoverridecommandlockouts                              

\usepackage{graphics} 
\usepackage{epsfig} 
\usepackage{mathrsfs} 
\usepackage{amsmath} 
\usepackage{amssymb}  
\usepackage{mathtools} 
\usepackage{amsfonts}
\usepackage{mathtools}
\usepackage{commath}
\usepackage{graphicx}
\usepackage[bookmarks=false]{hyperref}
\usepackage{comment}

\usepackage[noadjust]{cite}

\usepackage{xcolor}

\usepackage[font=small]{caption}
\usepackage{subcaption}

\usepackage{algorithm}
\newcommand{\RN}[1]{%
  \textup{\uppercase\expandafter{\romannumeral#1}}%
}
\usepackage[bookmarks=false]{hyperref}

\usepackage{kotex}

\usepackage{tikz}

\newcommand\copyrighttext{%
  \footnotesize \textcopyright 2021 IEEE.  Personal use of this material is permitted.  Permission from IEEE must be obtained for all other uses, in any current or future media, including reprinting/republishing this material for advertising or promotional purposes, creating new collective works, for resale or redistribution to servers or lists, or reuse of any copyrighted component of this work in other works.}
\newcommand\copyrightnotice{%
\begin{tikzpicture}[remember picture,overlay]
\node[anchor=south,yshift=10pt] at (current page.south) {\fbox{\parbox{\dimexpr\textwidth-\fboxsep-\fboxrule\relax}{\copyrighttext}}};
\end{tikzpicture}%
}

\title{\LARGE \bf
Real-Time Motion Planning of a Hydraulic Excavator \\ using Trajectory Optimization and Model Predictive Control
}

\author{Dongjae Lee$^{*}$, Inkyu Jang$^{*}$, Jeonghyun Byun, Hoseong Seo, and H. Jin Kim
\thanks{*The first two authors contributed equally to this work.}
\thanks{This research was supported by a research grant from Hyundai Construction Equipment.}%
\thanks{The authors are with the Department of Aerospace Engineering, Seoul National University (SNU), and Automation and Systems Research Institute (ASRI), Seoul 08826, South Korea {\tt\footnotesize \{ehdwo713, leplusbon, quswjdgus97, hosung37, hjinkim\}@snu.ac.kr}}%
}

\begin{document}

\maketitle
\copyrightnotice
\thispagestyle{empty}
\pagestyle{empty}

\begin{abstract}
Automation of excavation tasks requires real-time trajectory planning satisfying various constraints.
To guarantee both constraint feasibility and real-time trajectory re-plannability, we present an integrated framework for real-time optimization-based trajectory planning of a hydraulic excavator. The proposed framework is composed of two main modules: a global planner and a real-time local planner. The global planner computes the entire global trajectory considering excavation volume and energy minimization while the local counterpart tracks the global trajectory in a receding horizon manner, satisfying dynamic feasibility, physical constraints, and disturbance-awareness. We validate the proposed planning algorithm in a simulation environment where two types of operations are conducted in the presence of emulated disturbance from hydraulic friction and soil-bucket interaction: shallow and deep excavation. The optimized global trajectories are obtained in an order of a second, which is tracked by the local planner at faster than 30 Hz. To the best of our knowledge, this work presents the first real-time motion planning framework that satisfies constraints of a hydraulic excavator, such as force/torque, power, cylinder displacement, and flow rate limits.
\vspace{-0.2cm}
\end{abstract}

\section{Introduction}
Automation of heavy machinery, especially hydraulic excavators for its versatility, has been a steady research topic over a few decades. 
Various types of tasks, for example, soil digging \cite{son2020expert,sandzimier2020data}, rock moving \cite{sotiropoulos2020autonomous,mascaro2020towards}, and truck loading \cite{stentz1999robotic}, are studied  for excavator automation. To perform one of the most fundamental operation among such tasks, soil excavation, trajectory planning complying to \textit{operational constraints} (e.g. swept volume constraint during digging or bucket tip pose constraint during grading) and \textit{physical constraints} (e.g. actuator force/torque limit, pump flow rate limit, or power limit) is essential. To achieve such objective of planning a constrained reference trajectory, trajectory optimization has been widely applied \cite{yoo2010dynamics,kim2013dynamically,yang2020optimization,yang2020time}. However, since most existing works via trajectory optimization optimize the entire trajectory at once offline, they are vulnerable to instantaneous disturbance during operation. Furthermore, external disturbances to hydraulic excavators such as soil-bucket interaction and hydraulic friction are usually intractable and unignorable; accordingly, such unmodelable dynamics could result in sub-optimality and even constraint violation. 
Therefore, a new trajectory planning approach to consider constraints and compensate real-time disturbance is required.

\begin{figure}[t]
    \centering
    \includegraphics[width=0.8\linewidth]{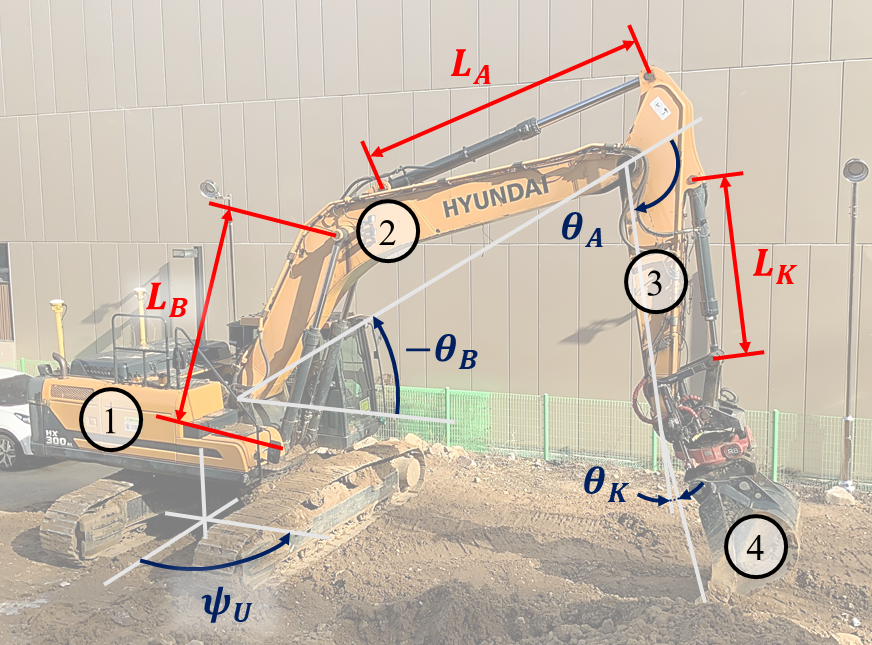}
    \caption{Excavator configuration of Hyundai HX300L. An excavator not in drive consists of four rotating rigid bodies: {\small \textcircled{\small{1}}} Cabin, {\small \textcircled{\small{2}}} Boom, {\small \textcircled{\small{3}}} Arm, and {\small \textcircled{\small{4}}} Bucket. The configuration can be defined either with four joint angles $\mathbf{q}_\theta=[\psi_U \ \theta_B \ \theta_A \ \theta_K]^\top$ or with one joint angle and three hydraulic cylinder displacements $\mathbf{q}_L=[\psi_U \ L_B \ L_A \ L_K]^\top$.}
    \label{fig:excavator_configuration}
    \vspace{-0.5cm}
\end{figure}

\subsection{Related work} 
We classify studies on excavator trajectory planning into three classes: 1) trajectory-optimization-based approach, 2) receding horizon optimization-based approach, and 3) non-optimization-based approach.

Given a specified cost function, constraints, and system dynamics, trajectory optimization generates an optimized trajectory in the defined state and input space which satisfies a set of constraints and system dynamics. Thanks to this property of explicit satisfaction of constraints, \cite{yoo2010dynamics,kim2013dynamically,yang2020time} successfully reflect the system dynamics with force/torque input, \textit{operational constraints}, and \textit{physical constraints} while minimizing overall effort or operation time. 
However, considering that optimization over the entire trajectory entails burdensome computation in general, the trajectories are usually parameterized using relatively simple kernel functions \cite{kim2013dynamically, yang2020optimization}.
Additionally, although soil model is employed to consider soil-bucket interaction forces \cite{kim2013dynamically,yang2020time}, the actual soil-bucket interaction may not be suitably addressed. 
There also exist studies regarding trajectory optimization of a new platform, e.g. a walking excavator \cite{jelavic2019whole,jelavic2020terrain}. Nevertheless, unlike other studies for conventional hydraulic excavator, they focus on planning and control of locomotion using newly added degrees of freedom (DoF) which is out of our scope. 

As a variation of trajectory optimization, receding horizon trajectory optimization or model predictive control (MPC), which optimizes trajectory in a receding horizon manner, is introduced \cite{bender2017modeling,wind2020trajectory}.  \cite{bender2017modeling} applies offset-free MPC to track the reference trajectory even in the presence of soil interaction disturbance while satisfying input box constraints. However, only decoupled dynamics, each with single input and no state constraints, are considered in MPC. In \cite{wind2020trajectory}, MPC is implemented for trajectory planning of an excavator, but system dynamics is modeled only with a double integrator, and the computation time seems to be insufficient for real-time application ($\approx$ 3.3 Hz).

Different from trajectory optimization, \cite{yang2019compact} computes a kinematically feasible reachability map of the bucket pose and linear movements offline. In \cite{son2020expert}, the authors apply dynamic motion primitives (DMP) to plan a trajectory which emulates experts' operation while being robust to various soil conditions. However, since these approaches cannot explicitly consider constraints, violation of constraints, especially \textit{physical constraints}, can occur. To handle such physical constraints, studies with a walking excavator \cite{jud2017planning,jud2019autonomous} present a hierarchical optimization-based controller. Nevertheless, feasibility cannot be always guaranteed due to the imposed hierarchy among constraints, and nonlinear constraints such as power constraint may not be directly considered in the optimization. In \cite{jud2017planning}, additional force controller tracking a force reference instead of a position reference is implemented in simulation to adjust to unknown soil conditions. 

\subsection{Algorithm overview}
\begin{figure}
    \centering
    \includegraphics[width = 1\linewidth]{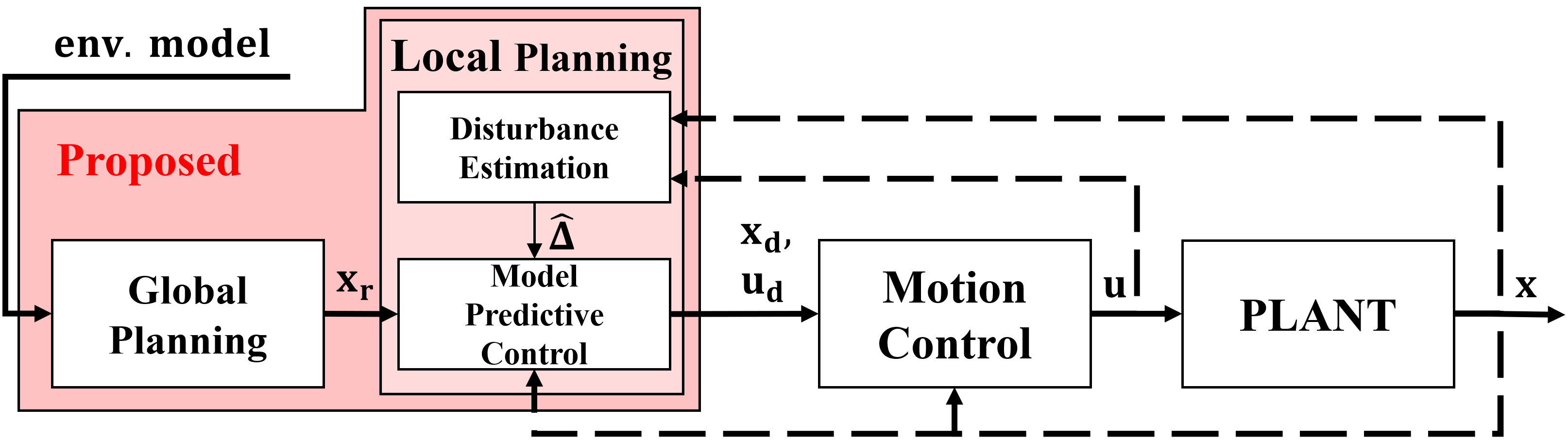}
    \caption{Flow chart for algorithm overview. Feedback signals are drawn with dashed lines.}
    \label{fig:flow_chart}
    \vspace{-0.5cm}
\end{figure}

To address both \textit{operational constraints} and \textit{physical constraints} in the presence of external disturbance during excavation, we present an integrated planner consisting of an online global planner and a real-time local planner as in \autoref{fig:flow_chart}. 
Considering that the global planning does not require any measurement feedback, the integrated planner can be executed in real-time if the global planner is computed within the trajectory run time.

In the global planning phase, a small-sized nonlinear optimization problem with a reduced number of constraints is solved. The obtained global trajectory is tracked by the model-predictive-control-based local planner, which plans high-fidelity local trajectories. The local planner runs alongside with a disturbance estimator, and takes the estimation results into consideration.
To the best of our knowledge, this work is the first real-time trajectory planning framework for autonomous excavation with guarantee of dynamic feasibility, physical constraint satisfaction, and disturbance-awareness. We summarize our main contributions as follows:
\begin{itemize}
    \item Integration of global planner and local planner for considering both operational and physical constraints.
    \item Formulation of the global trajectory optimization problem of the reduced dimension, which permits computation time shorter than a few seconds.
    \item Real-time constrained trajectory generation considering external disturbance.
\end{itemize}

\section{Equations of Motion}
\subsection{Kinematics}
Let $\mathscr{F}_B$ and $\mathscr{F}_C$ be the base frame and cabin frame where the base frame's relative orientation is defined by the ground slope, and the orientation of $\mathscr{F}_C$ differs from that of $\mathscr{F}_B$ by the amount of swing rotation. The origin and $x,z$-axis of $\mathscr{F}_C$ are illustrated in \autoref{fig:phase_description}. The configuration of the mechanical system of a hydraulic excavator can be either defined with $\mathbf{q}_{\theta}=[\psi_U \ \theta_B \ \theta_A \ \theta_K]^\top \in \mathbb{R}^4$ or $\mathbf{q}_L = [\psi_U \ L_B \ L_A \ L_K]^\top \in \mathbb{R}^4$ where $\psi_U,\theta_B,\theta_A,\theta_K$ are swing angle, rotation angles of boom, arm, and bucket, and $L_B,L_A,L_K$ are cylinder displacements of boom, arm, and bucket as shown in \autoref{fig:excavator_configuration}. The nonlinear, bijective\footnote{This is true within a physically feasible region, and the region is considered in the local planning as a cylinder displacement constraint.} mapping between $\mathbf{q}_\theta$ and $\mathbf{q}_L$ can be derived by solving closed-chain kinematics, denoted by
\begin{equation}
\begin{aligned}
    \mathbf{q}_\theta &= g_L(\mathbf{q}_L), &\ \mathbf{q}_L &= g_\theta(\mathbf{q}_\theta), \\
    \dot{\mathbf{q}}_\theta &= J_L(\mathbf{q}_L) \dot{\mathbf{q}}_L, &\ \dot{\mathbf{q}}_L &= J_\theta(\mathbf{q}_\theta) \dot{\mathbf{q}}_\theta,
\end{aligned}
\label{eq: kinematics-L to theta}
\end{equation}
where $J_L, J_\theta$ are corresponding Jacobian matrices. 

In some parts of this paper, the $\mathbf{q}_L$-configuration with fixed $\psi_U$ is used. For the sake of notational simplicity, we denote this subconfiguration by $\mathbf{L}=[L_B \ L_A \ L_K]^\top \in \mathbb{R}^3$. We also denote the pose of the bucket tip with respect to the $\mathscr{F}_C$ frame by $\mathbf{E} = [E_x \ E_z \ \theta]^\top \in \mathbb{R}^3$ where $E_x, E_z$ are the bucket tip's $x$, $z$-directional positions, and $\theta$ is its orientation, i.e., $\theta = \theta_B + \theta_A + \theta_K$. The components of $\mathbf{E}$ can be computed from $\mathbf{L}$ using the forward kinematics. We write them as $\mathbf{E} = [E_x(\mathbf{L}) \  E_z(\mathbf{L}) \  \theta(\mathbf{L})]^\top$.

\subsection{Dynamics}
By defining the system state $\mathbf{x}_\theta=[\mathbf{q}_\theta^\top \  \dot{\mathbf{q}}_\theta^\top]^\top\in \mathbb{R}^8$ and control input $\mathbf{u}=[T_U \ F_B \ F_A \ F_K]^\top\in \mathbb{R}^4$ where $T_U, \ F_B, \ F_A, \ F_K$ are swing torque, hydraulic forces of boom, arm, and bucket cylinder, the system dynamics can be derived with the Euler-Lagrange equation:
\begin{equation} \label{eq: dynamics}
    \dot{\mathbf{x}}_\theta = \left[\begin{matrix}
                        \dot{\mathbf{q}}_\theta \\
                        \ddot{\mathbf{q}}_\theta
                        \end{matrix}\right]
                    = \left[\begin{matrix}
                        \dot{\mathbf{q}}_\theta \\
                        M_\theta^{-1}(-h_\theta + J_\theta^\top \mathbf{u} + \Delta)
                        \end{matrix}\right]
                    \eqqcolon f_\theta(\mathbf{x}_\theta,\mathbf{u},\Delta)
\end{equation}
where $M_\theta(\mathbf{q}_\theta) \in \mathbb{R}^{4 \times 4}$, $h_\theta(\mathbf{q}_\theta,\dot{\mathbf{q}}_\theta) \in \mathbb{R}^4$ are the mass matrix and combined term for the Coriolis-centrifugal and gravitational effects. $\Delta \in \mathbb{R}^4$ is a lumped disturbance describing unmodeled dynamics, e.g. friction in hydraulic cylinders, soil-bucket interaction, and model uncertainties.

As explained in section \ref{sec: global planning} and \ref{sec: local planning}, since several state constraints considered in global and local planning can be expressed as linear constraints with respect to the hydraulic cylinder's motion, we further modify the system dynamics (\ref{eq: dynamics}) with the newly defined system state $\mathbf{x}_L = [\mathbf{q}_L^\top \ \dot{\mathbf{q}}_L^\top]^\top$. The equations of motion (\ref{eq: dynamics}) can be rewritten with respect to $\mathbf{x}_L$ and $\mathbf{u}$ as follows:
\begin{equation} \label{eq: dynamics_L}
    \dot{\mathbf{x}}_L = \left[\begin{matrix}
                        \dot{\mathbf{q}}_L \\
                        \ddot{\mathbf{q}}_L
                        \end{matrix}\right]
                    = \left[\begin{matrix}
                        \dot{\mathbf{q}}_L \\
                        M_L^{-1}(-h_L + \mathbf{u} + \Delta_L)
                        \end{matrix}\right]
                    \eqqcolon f_L(\mathbf{x}_L,\mathbf{u},\Delta_L)
\end{equation}
where $h_L(\mathbf{q}_L,\dot{\mathbf{q}}_L) = J_L^\top(\tilde{M}_L \dot{J}_L \dot{\mathbf{q}}_L + \tilde{h}_L)$, $\Delta_L=J_L^\top \Delta$, and $M_L(\mathbf{q}_L) = J_L^\top \tilde{M}_L J_L$. Here, by inserting the results of kinematics (\ref{eq: kinematics-L to theta}) into the $M_\theta(\mathbf{q}_\theta)$ and $h_\theta(\mathbf{q}_\theta,\dot{\mathbf{q}}_\theta)$, $\tilde{M}_L$ and $\tilde{h}_L$ are defined as
$\tilde{M}_L(\mathbf{q}_L) \coloneqq M_\theta(g_L(\mathbf{q}_L))$ and $\tilde{h}_L(\mathbf{q}_L,\dot{\mathbf{q}}_L) \coloneqq h_\theta(g_L(\mathbf{q}_L),J_L(\mathbf{q}_L)\dot{\mathbf{q}}_L)$.

\section{Global Planning}
\label{sec: global planning}
\begin{figure}
    \centering
    \includegraphics[width=0.3\textwidth]{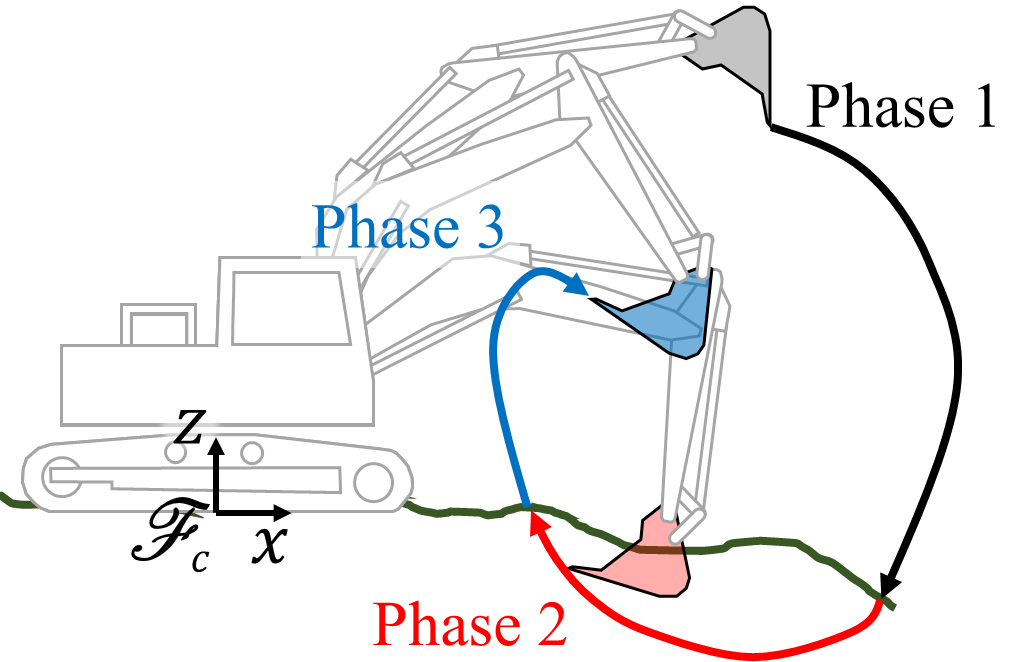}
    \caption{Illustration of the trace of the bucket tip during each phase of a single digging task. The origin of the base frame $\mathscr{F}_C$, which is used for the phase 2 global planning, is also depicted.}
    \label{fig:phase_description}
    \vspace{-0.5cm}
\end{figure}

A single excavation task of an excavator consists of three phases as follows:
\begin{itemize}
    \item \textbf{Phase 1}: The excavator tip moves from its initial position to the point where it enters the earth surface.
    \item \textbf{Phase 2}: The bucket tip cuts through the soil.
    \item \textbf{Phase 3}: The bucket conveys removed soil to the goal position, e.g., right above the dump box of a truck, without spilling the soil.
\end{itemize}
Details regarding each phase are described in \autoref{fig:phase_description}. Between consecutive phases, to guarantee continuity of control, configuration and its derivative must be matched.

\subsection{Global planning for phase 2}
In this subsection, we set up the trajectory optimization problem for phase 2. For phase 2, we constrain the cabin part of the excavator from moving, in order to prevent the side plates of the bucket from pushing against the soil. 
The goal of phase 2 global planning is to minimize the length travelled by the hydraulic cylinders during the excavation task.
Unlike phases 1 and 3, the excavator faces large and unpredictable external disturbance which originates from the direct contact between the earth and the bucket. 
Traditional torque- or force-minimizing strategies are therefore not suitable, since their optimality would be degraded due to external forces.
The travelled distance of the hydraulic cylinders should preferably serve as a cost measure, as the earth shall always resist the bucket's motion while in phase 2.
At the same time, we should prefer large excavation volume, because digging more volume will naturally reduce the number of repetitive excavation tasks needed to complete a mission.
To take the aforementioned considerations into account, we first discretize the phase 2 trajectory using $n_2 + 1$ waypoints, $\mathbf{L}_0$, $\mathbf{L}_1$, $\cdots$, $\mathbf{L}_{n_2}$ and propose the following cost function:
\begin{equation}
    J_{\mathrm{Phase 2}} = w_{{d}} \sum_{k=1}^{n_2} \norm{\mathbf{L}_k - \mathbf{L}_{k - 1}}_W^2 - w_{{v}} V,
\end{equation}
where $V$ is the volume of the removed soil, $w_{{d}} > 0$ and $w_{{v}} > 0$ are weights associated with the distance travelled by the cylinder and $V$, respectively. $W$ is a positive-definite weight matrix, where $\lVert x \rVert_Y$ ($Y>0$) denotes a weighted 2-norm of a vector $x$ with a matrix $Y$.

\begin{figure}
    \centering
    \includegraphics[width=0.25\textwidth]{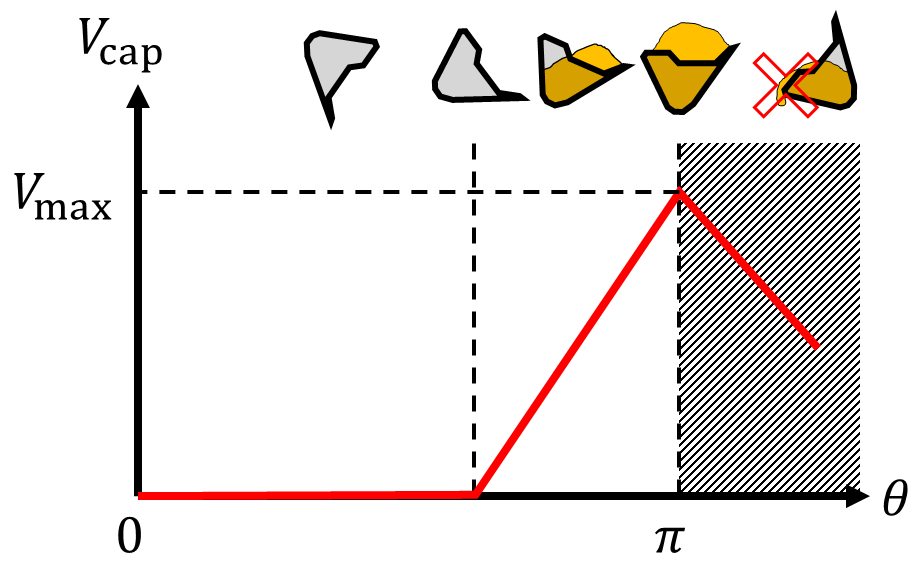}
    \caption{The capacity of the bucket $V_{\mathrm{cap}}$ with respect to its rotation angle $\theta$. If $\theta \geq \pi$, the bucket might spill its contents in the wrist direction, which is forbidden. The bucket is emptied when $\theta$ is smaller than a threshold determined by the bucket's geometry. }
    \label{fig:bucket_capacity}
    \vspace{-0.5cm}
\end{figure}

\begin{figure}
    \centering
    \includegraphics[width=0.4\textwidth]{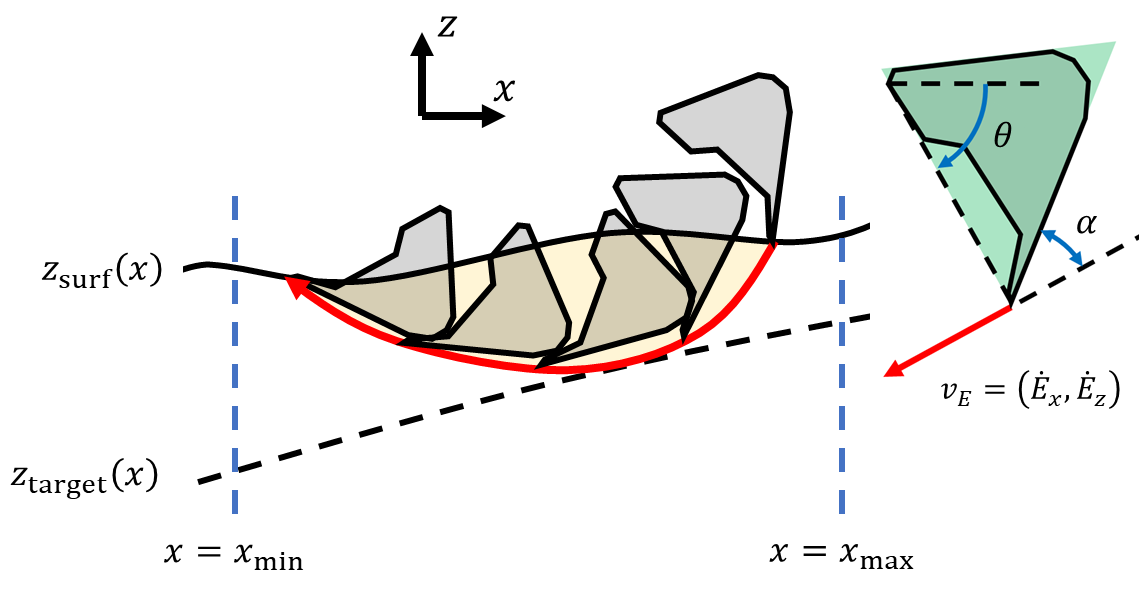}
    \caption{(left) A bucket trajectory that satisfies the constraints in phase 2. The current ground measurement and the target ground shape are denoted using solid and dashed black lines, respectively. (right) The bucket shape should not penetrate through the bucket tip trajectory. We use a triangular collision box to check this. The bucket rotation angle $\theta$ and the clearance angle $\alpha$ are denoted. $\alpha$ is mearsured with respect to the bucket tip velocity direction, $v_E=(\dot{E_x}, \dot{E_z})$, which is denoted by a red arrow.}
    \label{fig:bucket_constraint}
    \vspace{-0.5cm}
\end{figure}

In phase 2, earth surface $z_{\mathrm{surf}}$ and target shape $z_{\mathrm{target}}$ are described using polynomials over $x$, where $x$ and $z$ are given with respect to the frame $\mathscr{F}_C$ (\autoref{fig:phase_description}):
\begin{equation}
\begin{aligned}
    z_{\mathrm{surf}} &= a_0^{\mathrm{surf}} + a_1^{\mathrm{surf}}x + \cdots + a_{n_{\mathrm{surf}}}^{\mathrm{surf}}x^{n_{\mathrm{surf}}} \\
    z_{\mathrm{target}} &= a_0^{\mathrm{target}} + a_1^{\mathrm{target}}x + \cdots + a_{n_{\mathrm{target}}}^{\mathrm{target}}x^{n_{\mathrm{target}}}.
\end{aligned}
\end{equation}
Within the region of interest $x \in [x_{\mathrm{min}}, x_{\mathrm{max}}]$, the earth surface should be above the target, i.e., $z_{\mathrm{target}}(x) \leq z_{\mathrm{surf}}(x)$.
Now, we consider the following constraints for the bucket tip.
\begin{enumerate}
    \item The bucket tip should always be located within the region of interest, above the target surface and below the earth. At initial and final conditions, it must be located on the earth surface.
    \item The rotation angle $\theta$ of the bucket tip must be monotonically increasing; otherwise, the bucket plate pushes the dirt out of the bucket, which is energy-inefficient.
    \item The bucket must maintain positive clearance angle ($\alpha > 0$) while digging. This ensures that the bucket parts other than the tip do not push against the ground.
    \item In the similar sense, the whole bucket body must move above the tip path.
    \item The bucket rotation $\theta$ must be smaller than $\pi$: otherwise, the excavator will spill the dirt onto its wrist, which will escalate the risk of equipment malfunctioning. (\autoref{fig:bucket_capacity})
\end{enumerate}
The mentioned constraints are explained in \autoref{fig:bucket_constraint}.

Basically, the volume of the removed dirt is geometrically determined using the \textit{swept volume} between the earth surface and the bucket tip trajectory. However, if the swept volume is greater than the bucket capacity, the excavator shall spill the excess dirt, thus $V = \min{\left\{
        V_{\mathrm{swept}},
        V_{\mathrm{cap}}(\theta(\mathbf{L}_{n_2}))
    \right\}}$,
where
\begin{equation}
\begin{aligned}
    V_{\mathrm{swept}} =& \int_{x_{n_2}}^{x_0}{z_{\mathrm{surf}}(x)\; dx} \\
    {} - & {}  \frac{z_0 + 2z_1 + \cdots + 2z_{n_2 - 1} + z_{n_2}}{2n_2} \cdot \left(x_0 - x_{n_2}\right)
\end{aligned}
\end{equation}
is the (approximate) swept volume calculated using the trapezoidal integration rule,  $x_{i} = E_x (\mathbf{L}_{i})$, $z_{i} = E_z (\mathbf{L}_i)$, and $\theta_{i} = \theta(\mathbf{L}_i)$. The maximum capacity of the bucket, which is a function of $\theta_{n_2}$, is modeled as described in \autoref{fig:bucket_capacity}.

\subsection{Global planning for phases 1 and 3}
Unlike phase 2, in phases 1 and 3, the bucket travels a long distance while influenced by fewer constraints. 
Therefore, in order to reduce the computation burden, we parameterize the trajectory using a pre-selected set of bases. 
In this paper, Bernstein polynomials are used to parameterize the $\mathbf{q}_L$ trajectory.
Bernstein polynomial is written as
\begin{equation}
    p(s) = \sum_{k=0}^n \beta_k \cdot \binom{n}{k} s^k (1-s)^{n-k},
\end{equation}
where $s \in [0, 1]$.
Bernstein bases are used because the convex hull property of the bases allows to incorporate the box constraint for $\mathbf{q}_L$ and linear inequality constraints for $\dot{\mathbf{q}}_L$ easily. Furthermore, their time-derivatives can be easily calculated, so faster computation speed can be achieved. For phases 1 and 3, we minimize the force cost during travel.
\begin{equation}
\begin{aligned}
    \underset{\beta_k,\;T}{\mathrm{min.}} & \quad \int_{0}^{T} {\frac{1}{2}\mathbf{u}^\top W_u \mathbf{u}} \; dt \\
    \mathrm{s.t.} & \quad \mathbf{q}_L(t) = \sum_{k=0}^n {\beta_k \binom{n}{k} \left(\frac{t}{T}\right)^k \left(1-\frac{t}{T}\right)^{n-k}} \\
    & \quad {\mathbf{q}}_L^l \leq \beta_k \leq {\mathbf{q}}_L^u \quad \forall 0\leq k \leq n \\
    & \quad {\dot{\mathbf{q}}}_L^l \leq \frac{n\cdot (\beta_{k+1}-\beta_{k})}{T} \leq {\dot{\mathbf{q}}}_L^u 
    \quad \forall 0 \leq k \leq n-1\\
    & \quad \mathbf{q}_L(0) = \mathbf{q}_{L\cdot 0},\; \dot{\mathbf{q}}_L(0) = \dot{\mathbf{q}}_{L\cdot 0} \\
    & \quad \mathbf{q}_L(T) = \mathbf{q}_{L\cdot T},\; \dot{\mathbf{q}}_L(T) = \dot{\mathbf{q}}_{L\cdot T},
\end{aligned}
\end{equation}
where $\beta_k$ is the $k$-th control point for the Bernstein polynomial representation of $\mathbf{q}_L$, and $T$ is the time cost for complete traversal. Superscripts $(\cdot)^l$ and $(\cdot)^u$ denote lower and upper bounds for the state variables, respectively.
The boundary values $\mathbf{q}_{L\cdot 0}$, $\mathbf{q}_{L\cdot T}$, $\dot{\mathbf{q}}_{L\cdot 0}$, $\dot{\mathbf{q}}_{L\cdot T}$ are chosen such that the transition between any two consecutive phases is smooth. $W_u$ is a positive-definite weighting matrix.
The middle term in the third constraint is the control points for $\dot{\mathbf{q}}_L$ and was derived from the properties of Bernstein bases.
For phase 3, to keep the excavator from spilling the dirt, we introduce one additional constraint:
\begin{equation}
    \theta_0 \leq \theta(\beta_k) \leq \pi,
\end{equation}
where $\theta_0$ is the bucket rotation angle at which phase 2 ended. 

\section{Local Planning}
\label{sec: local planning}
This section describes how local reference trajectory is computed using MPC, which has been widely adopted in robotics as a real-time planner \cite{farshidian2017real,lee2020aerial,son2020real} thanks to its optimality and constraint-satisfying nature. The objective of this local receding-horizon planning is the following three: 1) physical constraint satisfaction, 2) disturbance-awareness, and 3) real-time computation. In addition to state constraints, actuation-related constraints are explicitly considered in the MPC problem formulation. Furthermore, to handle uncertainties in the dynamical model (\ref{eq: dynamics_L}), disturbance estimation is appended to the local planning for disturbance-awareness. Lastly, to guarantee real-time applicability of the proposed local planning, we re-formulate the MPC problem using feedback linearization as in \cite{van2006combined}. 
\vspace{-0.2cm}

\subsection{Constraints}
Similar to other existing researches for trajectory optimization of a hydraulic excavator \cite{yoo2010dynamics,kim2013dynamically}, we consider four types of physical constraints in the local planning: 1) actuation (force/torque) limit, 2) power limit, 3) cylinder displacement limit, and 4) pump flow rate limit. They are modeled as
\begin{subequations} \label{eq: constraints}
\begin{gather}
    \begin{split} \label{eq: input constraint}
    \mathbf{u}^l \leq \mathbf{u} \leq \mathbf{u}^u
    \end{split} \\
    \begin{split} \label{eq: power constraint}
    p = \mathbf{u}^\top \dot{\mathbf{q}}_L \leq p^u
    \end{split} \\
    \begin{split} \label{eq: length constraint}
    \mathbf{L}^l \leq \mathbf{L} \leq \mathbf{L}^u
    \end{split} \\
    \begin{split} \label{eq: flow rate constraint}
    f_i = A_i(\text{sgn}(\dot{\mathbf{q}}_L))^\top \dot{\mathbf{q}}_L^{\text{abs}} \leq f^u_i, \ i=1,2
    \end{split}
\end{gather}
\end{subequations}
where $\mathbf{u}^l,\mathbf{u}^u \in \mathbb{R}^4$ in (\ref{eq: input constraint}) are the lower and upper bounds of the control input $\mathbf{u}$, and $p^u\in \mathbb{R}$ in (\ref{eq: power constraint}) is the upper bound of the overall power generated by the hydraulic pump. It is required to impose this overall power limit because of the actuation mechanism of the hydraulic excavator where all four control inputs are powered by the same actuation source. $\mathbf{L}^l,\mathbf{L}^u \in \mathbb{R}^3$ in (\ref{eq: length constraint}) are the lower and upper bounds of the hydraulic cylinder displacement $\mathbf{L}$. Lastly, $f^u_i \in \mathbb{R}$ and $A_i = [A_{U,i} \ A_{B,i} \ A_{A,i} \ A_{K,i}]^\top \in \mathbb{R}^4$ in (\ref{eq: flow rate constraint}) are the upper bound of the pump flow rate and cross-section areas corresponding to swing motion, boom, arm, and bucket cylinder displacement, respectively. Because the cross-section area differs depending on the motion of a hydraulic cylinder, either contraction or expansion, $A_i$ is defined to include such effect and modeled as a function of sign of $\dot{\mathbf{q}}_L$ \cite{yoo2010dynamics}. $(\cdot)^{\text{abs}}$ is used to denote an element-wise absolute-valued vector mapped from a vector $(\cdot)$. To consider the flow rates of two hydraulic pumps, the subscript $i$ is used.             

\subsection{Disturbance estimation}
As mentioned in \cite{mayne2000constrained,li2011disturbance}, if disturbance is not properly considered in the system dynamics of MPC, constraints, especially state constraints, can be violated even if they were satisfied in the predicted trajectory computed with nominal dynamics.
To alleviate such feasibility issue, inspired by successful demonstrations of integration between model predictive control and disturbance estimation in robotics \cite{hentzen2019disturbance,kocer2019model}, we combined disturbance estimation and model predictive control for disturbance-awareness of the computed local reference trajectory.

We design a real-time disturbance estimation algorithm based on momentum-based disturbance estimation \cite{son2020expert,tomic2017external}. Considering the system dynamics based on the joint angles of the excavator (\ref{eq: dynamics}), the momentum-based disturbance estimation algorithm is formulated as follows:
\begin{equation} \label{eq: disturbance estimation}
\begin{aligned}
    \hat{\Delta} &= K_E \left(p - p(0) - \int_{0}^{t} \left(J_\theta^\top \mathbf{u} + C_\theta^\top \dot{\mathbf{q}}_\theta - G_\theta + \hat{\Delta}\right) \,d\tau \right) \\
    p &= M_\theta \dot{\mathbf{q}}_\theta
\end{aligned}
\end{equation}
where $C_\theta \in \mathbb{R}^{4 \times 4}, G_\theta \in \mathbb{R}^4$ are Coriolis matrix and gravity vector computed from the Euler-Lagrange equation. Utilizing the fact that $\dot{M}_\theta = C_\theta + C_\theta^\top$, the estimation algorithm can be analyzed to operate as a low pass filter of the actual disturbance $\Delta$ as $\dot{\hat{\Delta}} = K_E (\Delta - \hat{\Delta}).$ The disturbance estimate of the cylinder displacement-based dynamics (\ref{eq: dynamics_L}) then can be computed as $\hat{\Delta}_L = J_L^\top \hat{\Delta} \eqqcolon [\hat{\Delta}_U \ \hat{\Delta}_B \ \hat{\Delta}_A \ \hat{\Delta}_K]^\top.$

\subsection{Model predictive control}
Based on the system dynamics (\ref{eq: dynamics_L}) and constraints (\ref{eq: constraints}), the MPC problem is formulated as follows:
\begin{equation} \label{MPC problem}
\begin{aligned}
\underset{\mathbf{u}_k}{\textrm{min.}} \ & \ \lVert \mathbf{x}_N-\mathbf{x}_{r,N} \rVert_P^2 + \sum_{k=0}^{N-1} \lVert \mathbf{x}_k-\mathbf{x}_{r,k} \rVert_Q^2 + \lVert \mathbf{u}_k \rVert_R^2, \\
\textrm{s.t.} \ & \ \mathbf{x}_{k+1} = f_d(\mathbf{x}_{k},\mathbf{u}_{k},\hat{\Delta}_{k}) \\
& \ c_{k,i}(\mathbf{x}_{k},\mathbf{u}_{k}) \geq 0 \ \quad \forall 0 \leq i \leq m_k-1 \\
& \ c_{N,i}(\mathbf{x}_{k}) \geq 0 \ \quad \forall 0 \leq i \leq m_N-1
\end{aligned}
\end{equation}
where $N$ is the number of the prediction horizon, $f_d$ is the discretized dynamics of (\ref{eq: dynamics_L}), $c_{k,i}, c_{N,i}$ are the inequality constraints (\ref{eq: constraints}), and $m_k,m_N$ are the number of inequality constraints. $P$, $Q$, $R$ are positive-definite weighting matrices. Here, the subscript $L$ is omitted in the state $\mathbf{x}$, dynamics $f_d$, and disturbance estimate $\hat{\Delta}$ for brevity. A quadratic objective function is designed to allow the excavator to track the global reference trajectory $\mathbf{x}_r$ while minimizing the overall effort $\sum \lVert \mathbf{u} \rVert_R$. Lastly, similar to the widely adopted assumptions in papers on disturbance-aware model predictive control \cite{li2011disturbance,kocer2019model,hentzen2019disturbance}, we assume that the disturbance does not vary significantly over a short time horizon, i.e. $\hat{\Delta}_{k+1} = \hat{\Delta}_k$.

To solve the formulated MPC problem (\ref{MPC problem}), most optimal control solvers require partial derivatives, either numerical or analytic, of the dynamics (\ref{eq: dynamics_L}) and constraints (\ref{eq: constraints}) with respect to the state $\mathbf{x}$ and the input $\mathbf{u}$. However, due to the high nonlinearity of the system dynamics (\ref{eq: dynamics_L}), especially the matrix inverse of the mass matrix $M_L$ which inheres in the closed-chain kinematics, it is difficult to obtain an optimized solution in real-time. Therefore, we simplify the MPC problem by introducing a virtual input $\mathbf{v} \in \mathbb{R}^4$ and a feedback-linearization-based control law $\mathbf{u} = \text{FL}(\mathbf{x},\mathbf{v}) \coloneqq h_L + M_L \mathbf{v} - \hat{\Delta}_L$ through which the original nonlinear dynamics (\ref{eq: dynamics_L}) is transformed into a linear dynamics as 
\begin{equation}\label{eq: feedback linearization - linear dynamics}
\begin{gathered}
    \dot{\mathbf{x}} = 
    A_c \mathbf{x} + 
    B_c \mathbf{v} \\
    A_c = \left[\begin{matrix}
    0_{4\times4} & I_{4} \\
    0_{4\times4} & 0_{4\times4}
    \end{matrix}\right], 
    B_c = \left[\begin{matrix}
    0_{4\times4} \\ I_4
    \end{matrix}\right] 
\end{gathered}
\end{equation}
where $0_{4\times4}, I_4 \in \mathbb{R}^{4\times 4}$ are zero and identity matrices.
The feedback-linearization-based MPC problem is then formulated as follows:
\begin{equation} \label{FL-MPC problem}
\begin{aligned}
\underset{\mathbf{v}_k}{\textrm{min.}} \ & \ \lVert \mathbf{x}_N-\mathbf{x}_{r,N} \rVert_{\tilde{P}}^2 + \sum_{k=0}^{N-1} \lVert \mathbf{x}_k-\mathbf{x}_{r,k} \rVert_{\tilde{Q}}^2 + \lVert \mathbf{v}_k \rVert_{\tilde{R}}^2, \\
\textrm{s.t.} \ & \ \mathbf{x}_{k+1} = A_d \mathbf{x}_k + B_d \mathbf{v}_k \\
& \ c_{k,i}(\mathbf{x}_{k},\text{FL}(\mathbf{x}_{k},\mathbf{v}_{k})) \geq 0 \ \quad \forall 0 \leq i \leq m_k-1 \\
& \ c_{N,i}(\mathbf{x}_{k}) \geq 0 \ \quad \forall 0 \leq i \leq m_N-1
\end{aligned}
\end{equation}
where $\tilde{P}$, $\tilde{Q}$, $\tilde{R}$ are positive-definite weight matrices, and $A_d, B_d$ are constant matrices in the discretized dynamics, corresponding to $A_c$ and $B_c$ in the continuous counterpart (\ref{eq: feedback linearization - linear dynamics}).

\section{Simulation}
\subsection{Setup}
We validate the proposed planning algorithms in simulation. The overall flow chart during simulation is illustrated in \autoref{fig:flow_chart}. All modules in \autoref{fig:flow_chart} are individually implemented in Robot Operating System (ROS) using C++. Excavator dynamics simulator (PLANT in \autoref{fig:flow_chart}) is based on the derived system dynamics (\ref{eq: dynamics}) where the external disturbance $\Delta$ is modeled with a steady-state modified LuGre friction model \cite{tran2012modeling} for cylinder friction and modified Fundamental Earth-moving Equation (FEE) \cite{luengo1998modeling,kim2013dynamically} for soil-bucket interaction. Note that friction and soil-bucket interaction models are introduced only to construct a realistic simulation environment and are unknown to the planning algorithms. Parameters in the system dynamics and physical constraints are from the Hyundai HX300L model in \autoref{fig:excavator_configuration}.

As a motion controller, which computes the control input $\mathbf{u}$ based on the reference trajectory $\mathbf{x}_d$ generated from the local planning module, we adopt a feedback linearization-based PID controller: $\mathbf{u} = J_\theta^{-\top}(h_\theta - \hat{\Delta} + M_\theta \mathbf{u}_{\text{PID}})$. Applying the control input $\mathbf{u}$ to the system equation (\ref{eq: dynamics}), input-output stability of the closed-loop system can be derived \cite[Ch. 5]{khalil2002nonlinear} where the input is the estimation error of the external disturbance $\tilde{\Delta} = \Delta - \hat{\Delta}$ multiplied by the inverse of the mass matrix $M_\theta$, and the output is the state $\mathbf{x}$.

The global planner is implemented using ALGLIB \cite{bochkanov2013alglib}, an open-source nonlinear optimization solver. We use 9 Bernstein control points for phases 1 and 3, respectively, and 20 via points for phase 2.
Next, to solve the formulated model predictive control problem (\ref{FL-MPC problem}), we implement differential dynamic programming (DDP) with augmented Lagrangian method \cite{plancher2017constrained}, which is an algorithm widely exploited in various robotic application for real-time trajectory optimization \cite{son2020real,farshidian2017real,lee2020aerial}. It transforms the original constrained MPC problem into an unconstrained problem using the Lagrangian and quadratic penalty function, which is then solved with the conventional DDP algorithm. For MPC implementation, we employ the time horizon of 1000 ms, control discretization and integration rate of 20 ms. Except for the results of cylinder displacements $\mathbf{L}$, which are normalized with constant bias and constant normalization factors, the other results related to physical limits of Hyundai HX300L displayed in \autoref{fig:scenario1} and \autoref{fig:scenario2} are normalized with constant normalization factors, denoted with $\bar{(\cdot)}$. Each element of control inputs and the corresponding channel of disturbance estimates are normalized with the same factor for comparison.

To validate the proposed framework, we conducted two simulation scenarios on a desktop computer equipped with an Intel Core i7-9750 CPU at 2.6 GHz, by varying the target excavation depth: 1) shallow excavation and 2) deep excavation. 
As depicted in \autoref{fig:flow_chart}, we assume prior knowledge of the environment, i.e. the ground shape, which can be obtained either from onboard sensors like lidars and cameras, or from external sensors like total station in real experiment.

\subsection{Results}
\subsubsection{Scenario 1 - shallow excavation}
\begin{figure*}
    \centering
    \includegraphics[width=0.7\textwidth]{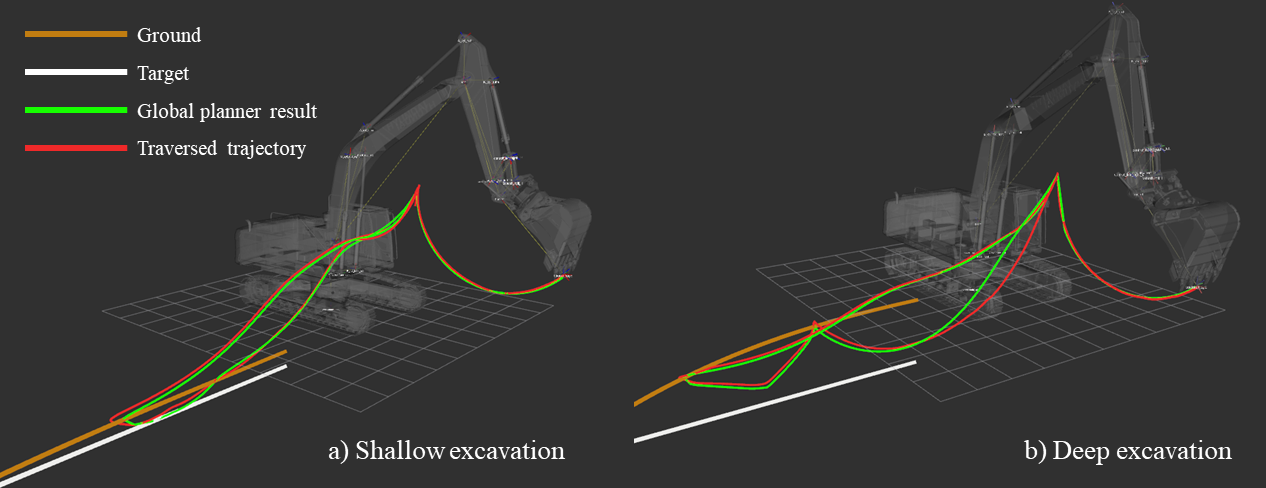}
    \caption{Global planning results for the two scenarios.}
    \label{fig:global_planning_result}
    \vspace{-0.45cm}
\end{figure*}

\begin{figure*}[t]
    \centering
    \begin{subfigure}[b]{0.47\linewidth}
        \centering
        \includegraphics[width = 1\linewidth]{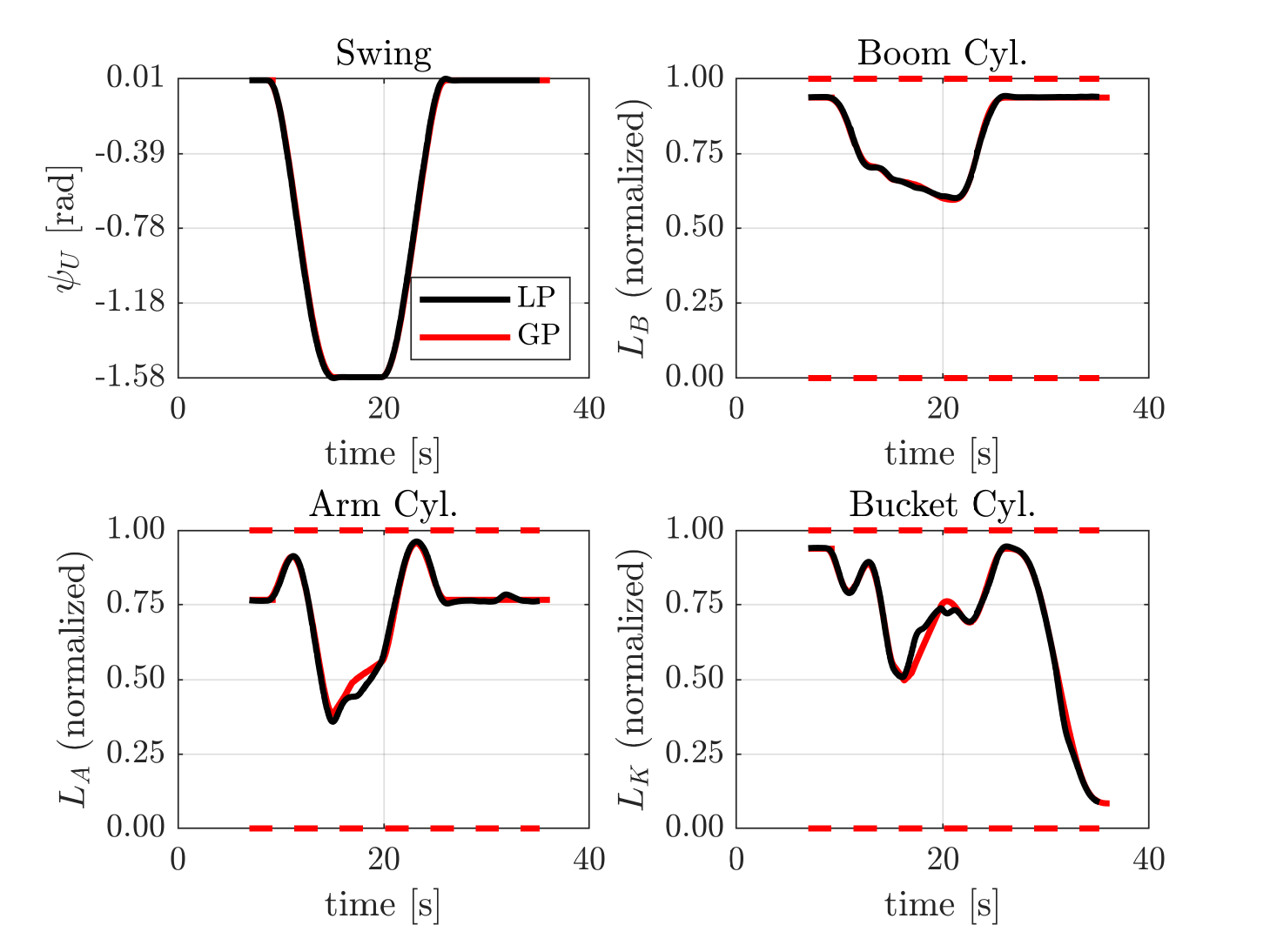}
        \caption{local planning result: global reference tracking}
        \label{fig:scenario1-local_planning_result-ref_track}
    \end{subfigure}
    \hfill
    \begin{subfigure}[b]{0.47\linewidth}
        \centering
        \includegraphics[width = 1\linewidth]{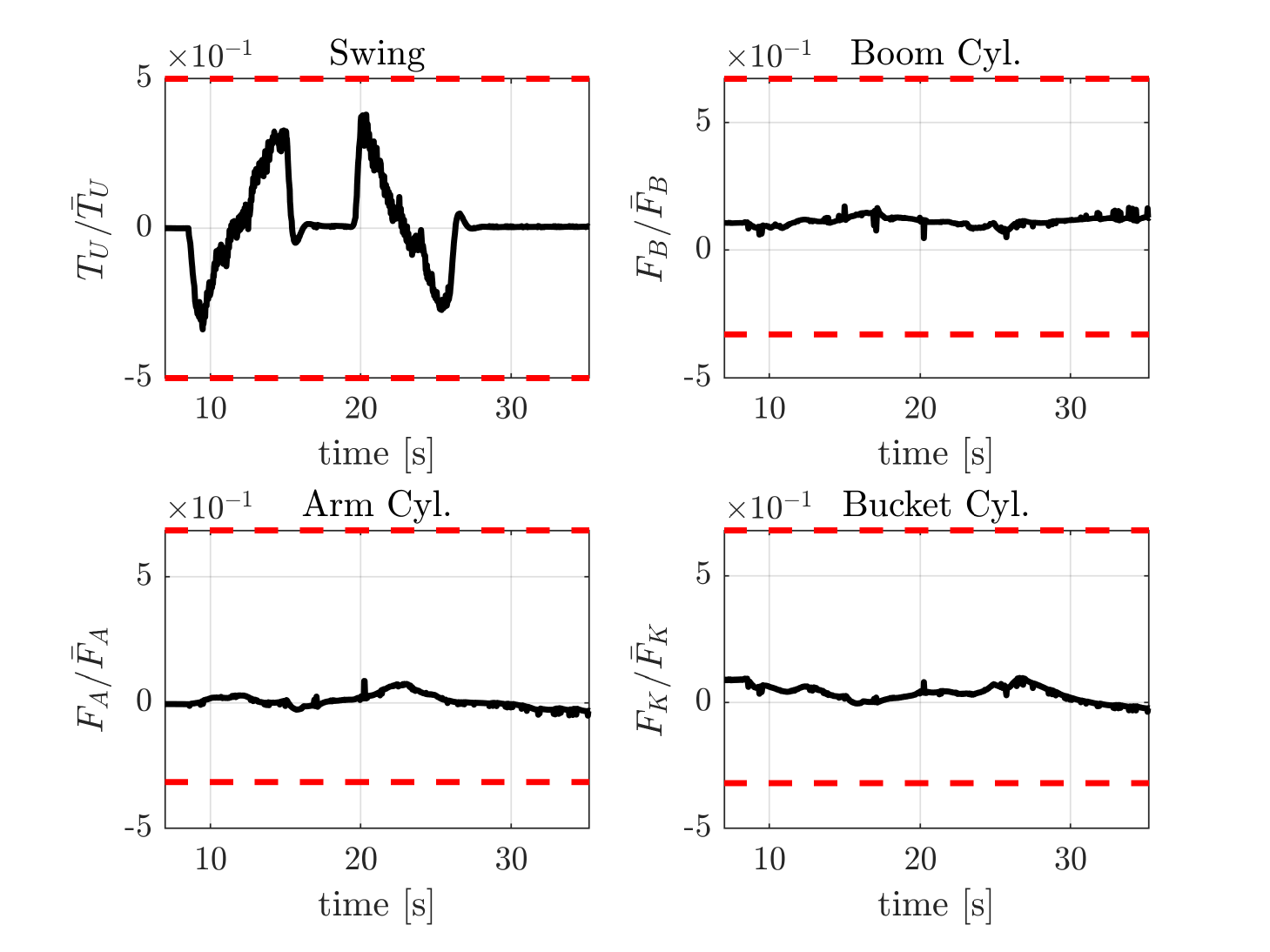}
        \caption{local planning result: normalized control input trajectory}
        \label{fig:scenario1-local_planning_result-input}
    \end{subfigure}
    \begin{subfigure}[b]{0.47\linewidth}
        \centering
        \includegraphics[width = 1\linewidth]{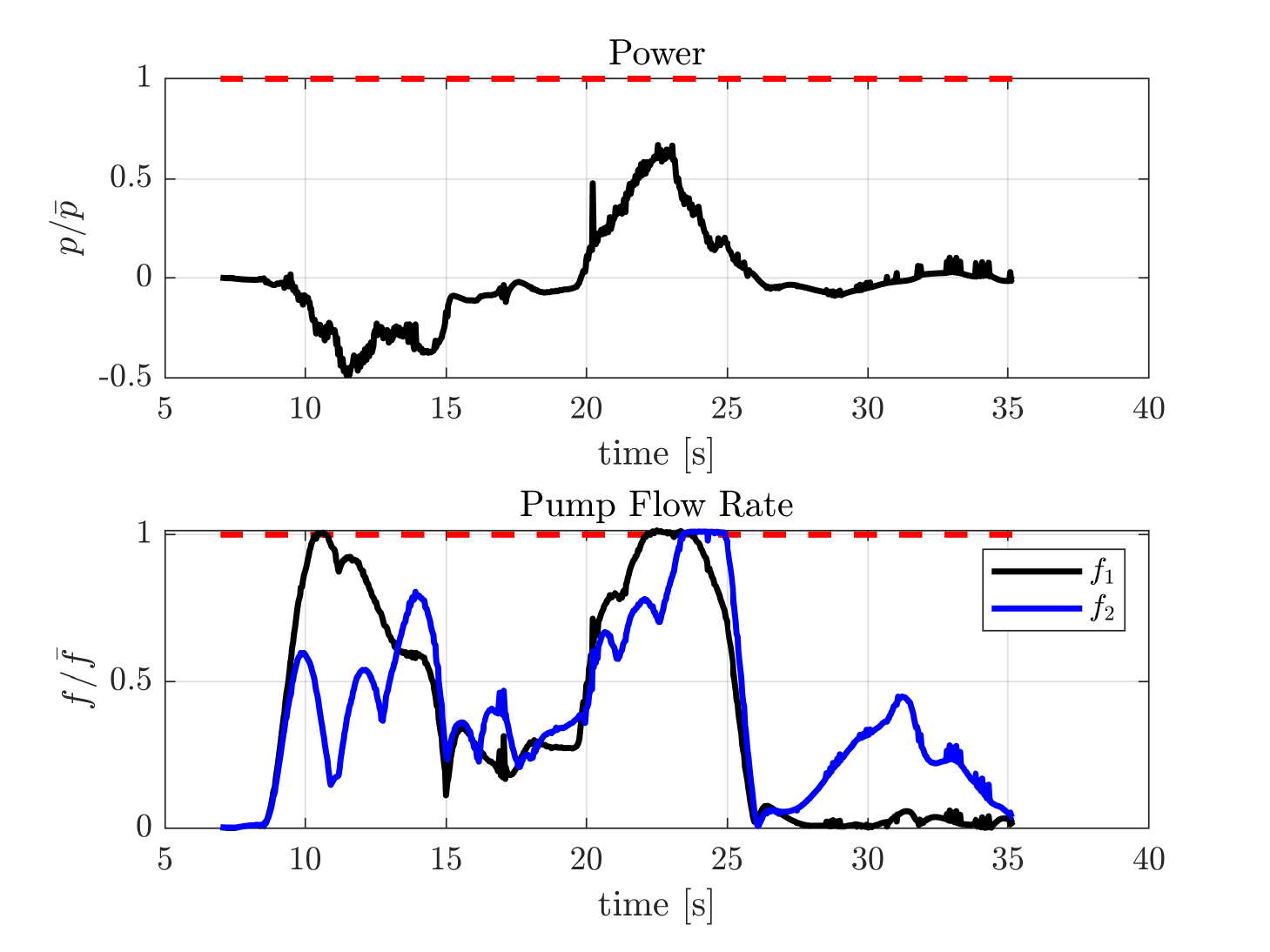}
        \caption{local planning result: normalized power and pump flow rate}
        \label{fig:scenario1-local_planning_result-const_sat}
    \end{subfigure}
    \hfill
    \begin{subfigure}[b]{0.47\linewidth}
        \centering
        \includegraphics[width = 1\linewidth]{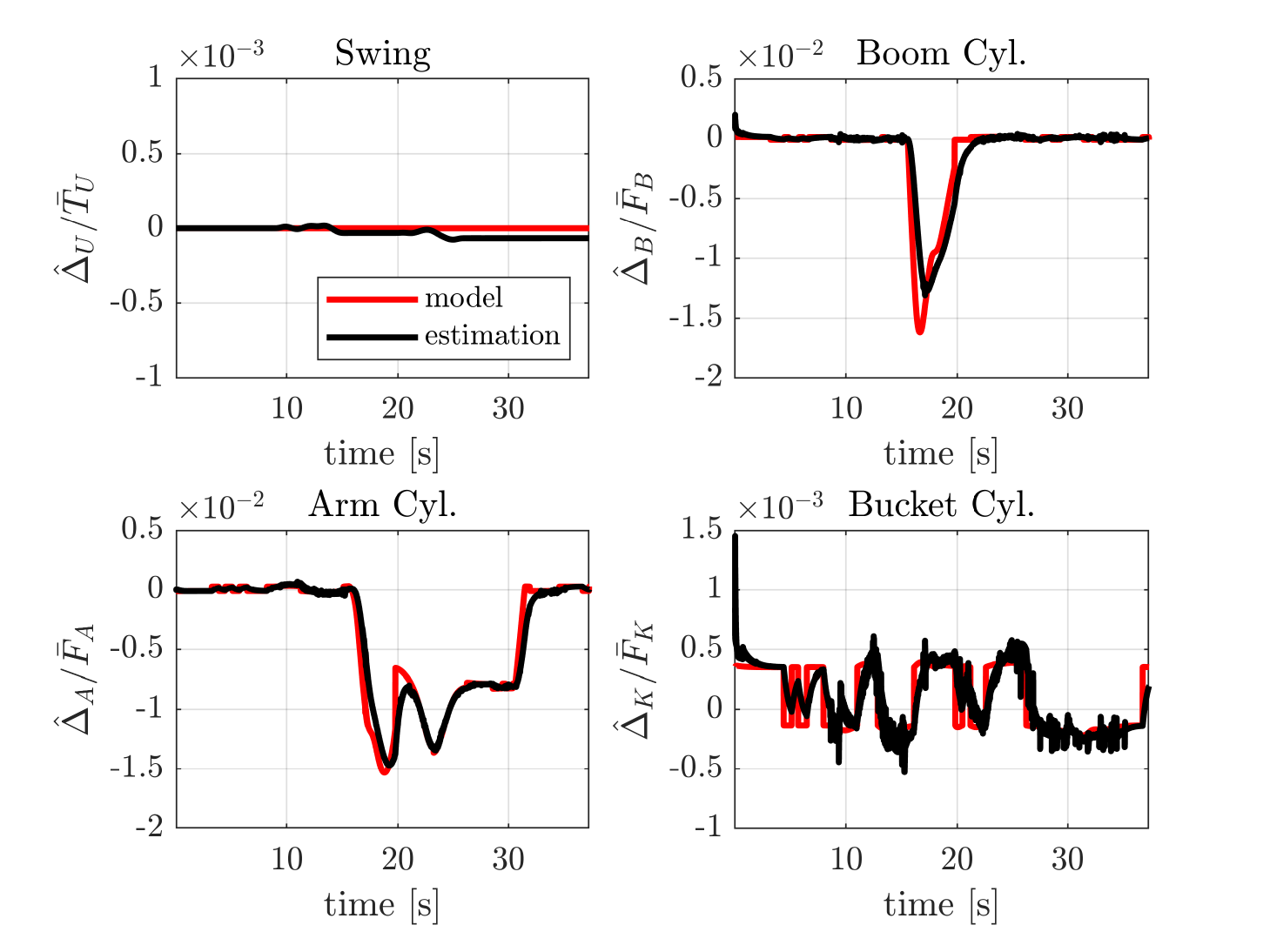}
        \caption{local planning result: normalized estimated disturbance}
        \label{fig:scenario1-local_planning_result-dist_est}
    \end{subfigure}
    \caption{Scenario 1 - shallow excavation. In the subfigures (a), (b), and (c), red dashed lines denote the corresponding physical limits. LP and GP in the subfigure (a) stand for the trajectory computed from local planning and that from global planning. Each variable with a bar in the subfigures (b), (c), (d) is the corresponding constant normalization factor.}
    \label{fig:scenario1}
    \vspace{-0.5cm}
\end{figure*}

\autoref{fig:global_planning_result}a and \autoref{fig:scenario1} describe the simulation result for shallow excavation. 
In the first scenario, not to penetrate through the target shape, the global planner plans a trajectory that stretches out in a relatively large region in phase 2. The bucket tip well tracks the target shape, and tip-grading-like motion is generated. The global trajectory was generated in 2.1 seconds.
As in \autoref{fig:scenario1-local_planning_result-ref_track} and \autoref{fig:scenario1-local_planning_result-const_sat}, even if the global trajectory is intentionally generated to exceed the physical limit of some variables that are not considered in the global planning, physical constraints are still satisfied during local planning phase. The control input trajectory computed from the local planning is displayed in \autoref{fig:scenario1-local_planning_result-input} where all the input limits are satisfied. In \autoref{fig:scenario1-local_planning_result-dist_est}, the hydraulic cylinder friction model can be clearly found at the bucket cylinder (bottom-left subfigure), whereas the effect by the friction model is found to be smaller at the boom cylinder (top-right subfigure) and the arm cylinder (bottom-left) due to larger disturbance from the soil-bucket interaction. Still, the disturbance estimate (black line) follows the modeled disturbance (red line) in all directions. 
The average computation time of the local planning is 21.7 ms ($\approx$ 46 Hz) which we believe is sufficient for real-time application.

\subsubsection{Scenario 2 - deep excavation}

\begin{figure*}[t]
    \centering
    \begin{subfigure}[b]{0.47\linewidth}
        \centering
        \includegraphics[width = 1\linewidth]{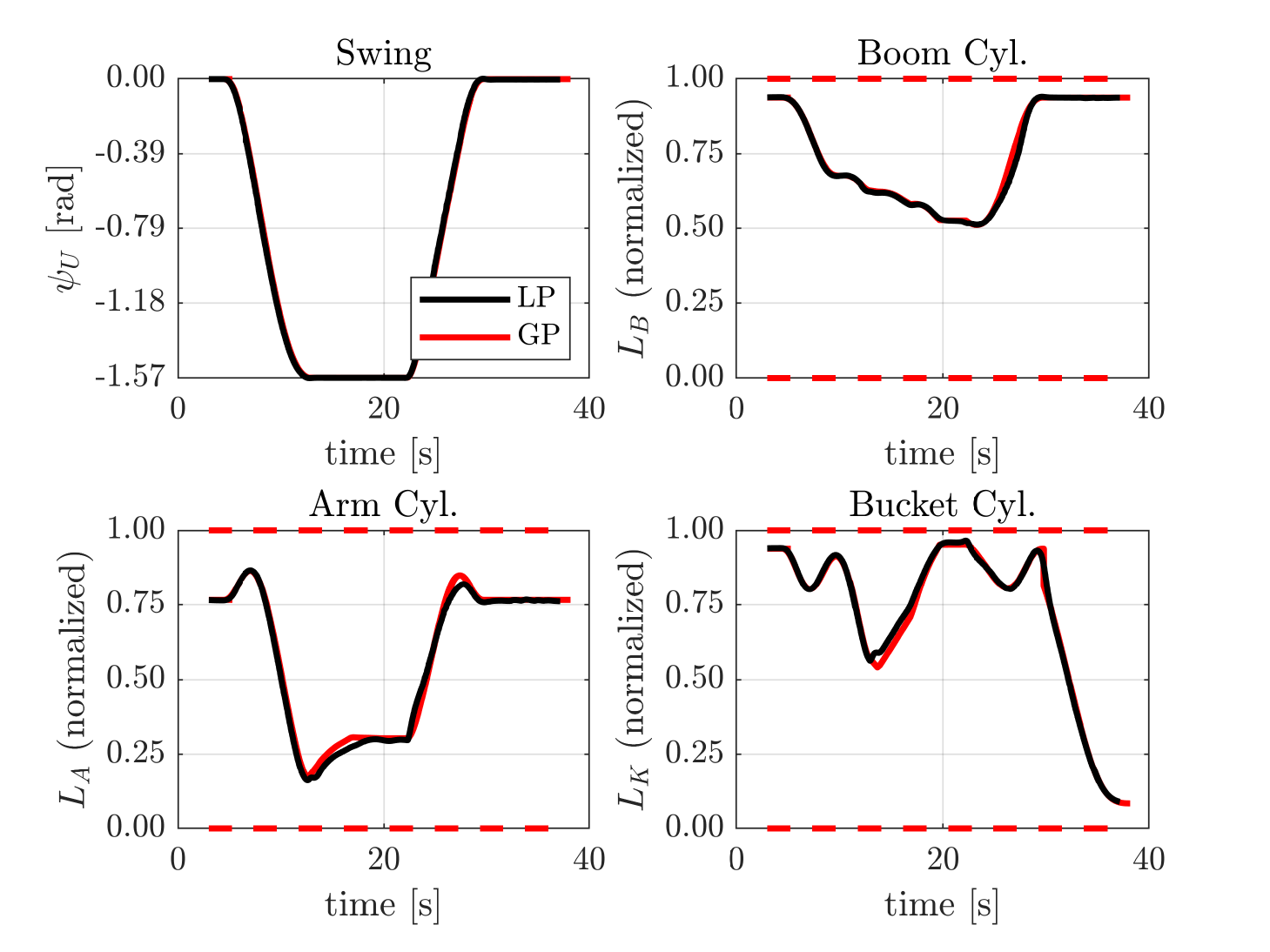}
        \caption{local planning result: global reference tracking}
        \label{fig:scenario2-local_planning_result-ref_track}
    \end{subfigure}
    \hfill
    \begin{subfigure}[b]{0.47\linewidth}
        \centering
        \includegraphics[width = 1\linewidth]{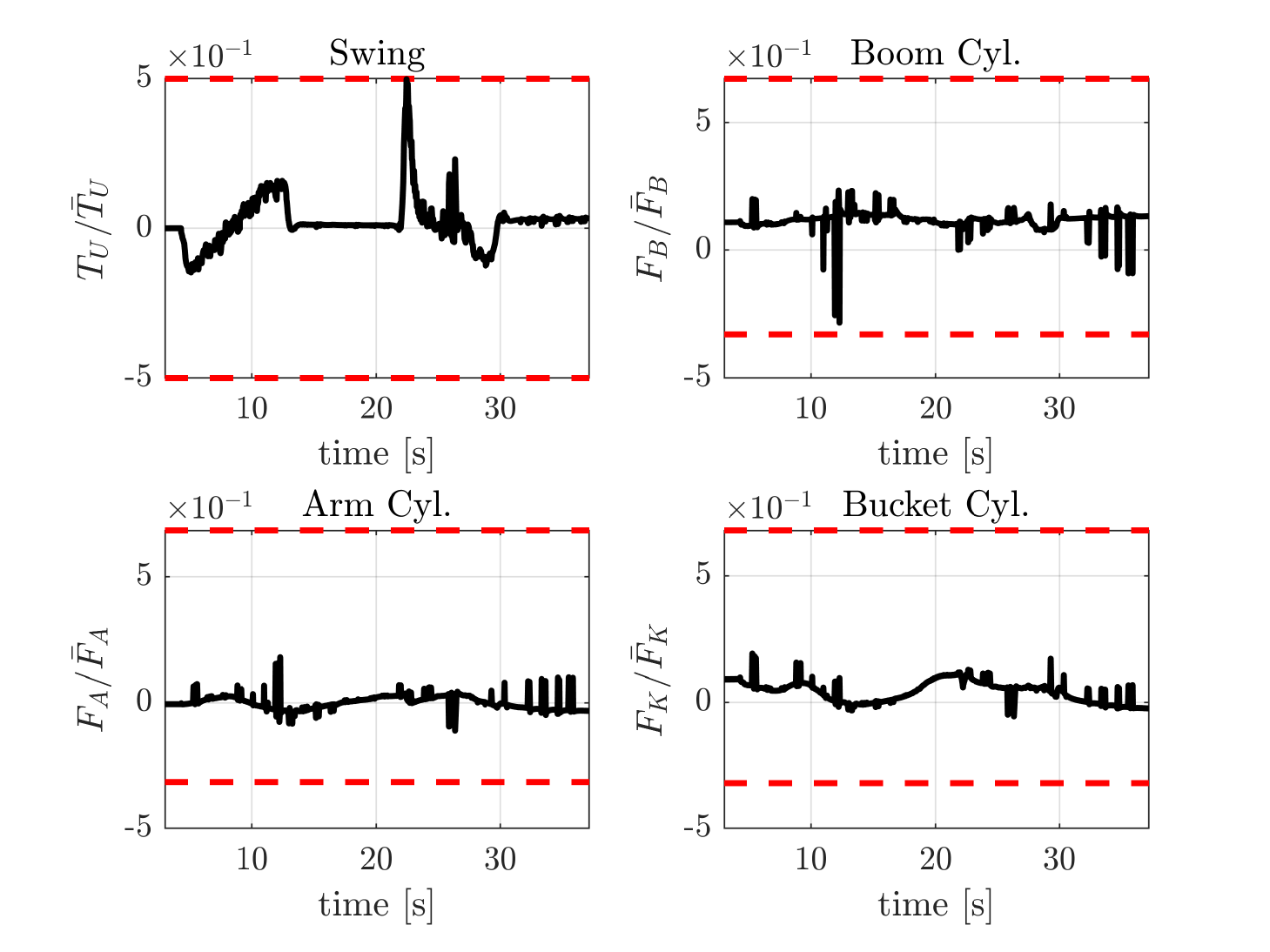}
        \caption{local planning result: normalized control input trajectory}
        \label{fig:scenario2-local_planning_result-input}
    \end{subfigure}
    \begin{subfigure}[b]{0.47\linewidth}
        \centering
        \includegraphics[width = 1\linewidth]{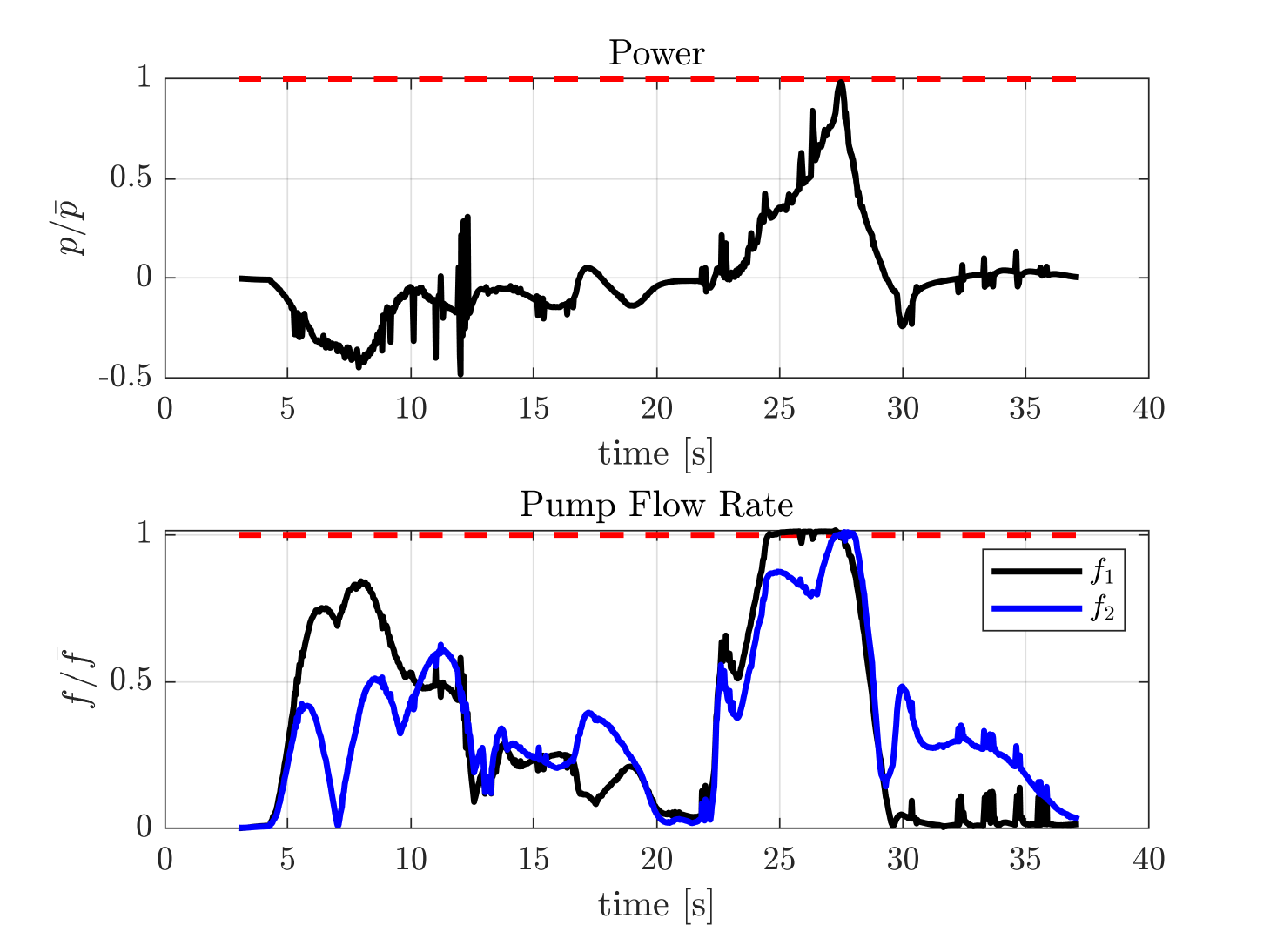}
        \caption{local planning result: normalized power and pump flow rate}
        \label{fig:scenario2-local_planning_result-const_sat}
    \end{subfigure}
    \hfill
    \begin{subfigure}[b]{0.47\linewidth}
        \centering
        \includegraphics[width = 1\linewidth]{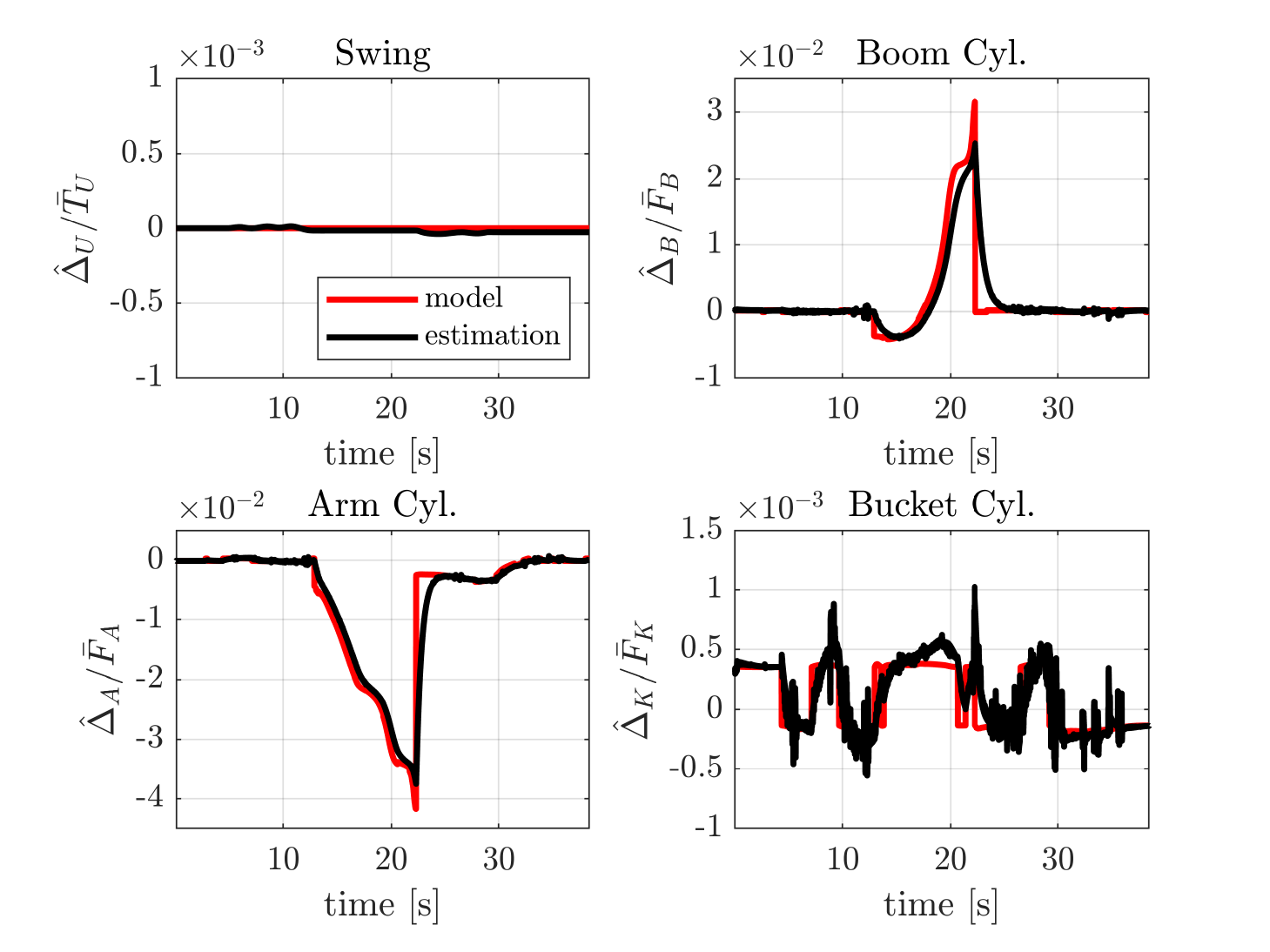}
        \caption{local planning result: normalized estimated disturbance}
        \label{fig:scenario2-local_planning_result-dist_est}
    \end{subfigure}
    \caption{Scenario 2 - deep excavation. In the subfigures (a), (b), and (c), red dashed lines denote the corresponding physical limits. LP and GP in the subfigure (a) stand for the trajectory computed from local planning and that from global planning. Each variable with a bar in the subfigures (b), (c), (d) is the corresponding constant normalization factor.}
    \label{fig:scenario2}
    \vspace{-0.5cm}
\end{figure*}

Simulation results of the second scenario can be found in \autoref{fig:scenario2} and \autoref{fig:global_planning_result}b. 
In contrast to the first scenario, if the target ground is located deep below the earth surface, the excavator tip does not aim to reach the target shape in phase 2. Instead, it maximizes the excavation volume by generating a triangle-shaped phase 2 trajectory. Since the phase 2 trajectory is constrained not to exceed the target shape, such deep target shape can be finally reached by repeating this target-shape-aware excavation-volume-maximizing strategy. The global trajectory was generated in 1.6 seconds.
As in the subfigure \autoref{fig:scenario2-local_planning_result-dist_est}, larger disturbance than that in the first scenario was applied due to deeper penetration of the bucket into the soil. Even in the presence of such larger disturbance, similar to the results of the first scenario, the global reference is well tracked by the local planning as can be found in the subfigure \autoref{fig:scenario2-local_planning_result-ref_track}. While tracking the global reference, the computed local trajectory is shown to satisfy all the physical constraints as in the subfigures \autoref{fig:scenario2-local_planning_result-ref_track},  \autoref{fig:scenario2-local_planning_result-input}, and \autoref{fig:scenario2-local_planning_result-const_sat}. It took 30.2 ms ($\approx$ 33 Hz) for local planning computation on average.

\section{Conclusion}
In this paper, we presented a real-time planning algorithm for a single digging task of a hydraulic excavator. The algorithm consists of a global and a local planner. In the global planning phase, the trajectory that maximizes excavation volume and minimizes energy cost under the \textit{operational constraints} is generated online using Bernstein parameterization. The high-fidelity local planner tracks the generated global trajectory while satisfying the \textit{physical constraints} in real-time.
The proposed algorithm was validated through a realistic simulation environment with two different operation objectives: deep and shallow soil excavation. The simulation results showed that the proposed online global planner generates an optimized bucket tip trajectory in a few seconds, and the local planner plans disturbance-aware, physically feasible receding horizon trajectories in real-time. 
Future work may include validation of the proposed algorithm in actual excavators.





\section*{ACKNOWLEDGMENT}
The authors would like to appreciate Young Bum Kim (Hyundai Construction Equipment) for his kind support in the simulation setup.


\end{document}